\documentclass[10pt, a4paper]{article}

\usepackage{lrec-coling2024} 
\usepackage{times}
\usepackage{latexsym}
\usepackage{amsmath}
\usepackage{algorithm}
\usepackage[noend]{algpseudocode}
\usepackage{graphicx}
\usepackage{subcaption}
\usepackage{booktabs}
\usepackage{longtable}
\usepackage{todonotes}
\usepackage[normalem]{ulem}
\usepackage{soul, color}
\usepackage{enumitem}
\usepackage{multirow}
\usepackage{multicol}
\usepackage{arydshln}
\usepackage{microtype}
\usepackage{setspace}
\usepackage{wrapfig}

\def\modelname{\texttt{EROS}}
\def\dataset{{\texttt {PD-Sum}}}
\def\nermodelname{{\texttt {EEPD}}}
\def\ztitle{EROS: Entity-Driven Controlled Policy Document Summarization}
\title{\ztitle}

\name{Joykirat Singh$^*$\thanks{$^*$First two authors contributed equally}, Sehban Fazili$^*$, Rohan Jain, Md Shad Akhtar} 

\address{Indraprastha Institute of Information Technology Delhi (IIIT Delhi), India. \\
         \{\tt joykirat19166, sehban21143, rohan19095, shad.akhtar\}@iiitd.ac.in\\}

\abstract{
Privacy policy documents have a crucial role in educating individuals about the collection, usage, and protection of users' personal data by organizations. However, they are notorious for their lengthy, complex, and convoluted language especially involving privacy-related entities. 
Hence, they pose a significant challenge to users who attempt to comprehend organization's data usage policy. In this paper, we propose to enhance the interpretability and readability of policy documents by using controlled abstractive summarization -- we enforce the generated summaries to include critical privacy-related entities (e.g., \textit{data} and \textit{medium}) and organization's rationale (e.g., \textit{target} and \textit{reason}) in collecting those entities. To achieve this, we develop \dataset, a policy-document summarization dataset with marked privacy-related entity labels. Our proposed model, \modelname, identifies critical entities through a span-based entity extraction model and employs them to control the information content of the summaries using proximal policy optimization (PPO). 
Comparison shows encouraging improvement over various baselines. Furthermore, we furnish qualitative and human evaluations to establish the efficacy of \modelname.
 \\ \newline  }

\begin{document}

\maketitleabstract

\section{Introduction}


In today's Internet era, access to information has never been so convenient. Every day, an overwhelming amount of users are exploring the Internet horizon for entertainment or business purposes. Realizing an excellent opportunity to increase their customer base, many organizations offer their products or services in a convenient online setting. In majority of the cases, customers are required to sign up to acquire the services on offer and while doing so, they have to agree to the term-and-conditions (T\&C) or policies of the service providers.

A privacy policy is a crucial component of any organization that allows it to legally collect, process, store, and/or distribute personal information. It outlines how an organization will handle personal data and how it will comply with applicable data protection laws and regulations. Little that they know, on many occasions, customers are, advertently or inadvertently, granting full access to their sensitive and private data (e.g., name, contact information, location, etc.) to the service providers without reading or understanding the privacy policy document. Moreover, some companies collect data with distributional rights as well and make a fortune by selling user's data to third parties without their realization but with their inadvertent consent\footnote{\href{https://www.searchenginejournal.com/brave-browser-under-fire-for-alleged-sale-of-copyrighted-data/491854/}{Brave under fire for alleged sale of copyrighted data}}$^\text{,}$\footnote{\href{https://www.bbc.com/news/technology-61606476}{Twitter fined 150m in US for selling user's data}}. The primary reason for such ignorance on the user's part is their busy and packed schedule as well as lengthy and technical/legal language, which are usually difficult to comprehend by a common user.   




\begin{figure*}
    \centering
    \subfloat[\label{tab:entity-definition}Entity labels with their explanation]{
        \resizebox{0.55\textwidth}{!}{
            \renewcommand*{\arraystretch}{1.2}
            \begin{tabular}{l|l} 
            \toprule
             \textbf{Entity} & \textbf{Explanation} \\  \hline
             \textcolor{purple}{Data Compulsory} & Data which is compulsory for the source to enter \\  \hdashline
             \textcolor{olive}{Data Optional} & Data in which the source has the option to provide to the target  \\  \hdashline
             \textcolor{brown}{Data Others} & Information that belongs to the source \\  \hline
             
             \textcolor{red}{Source Direct} & The entity that directly provides the data to the target \\  \hdashline
             Source Indirect & The entity that indirectly provides the data to the target \\ \hline
             \textcolor{teal}{Target Direct} & The entity to which the source directly provides the data \\  \hdashline
             \textcolor{orange}{Target Indirect} & The entity to which the source indirectly provides the data \\  \hline
             \textcolor{violet}{Reason} & Reason for why the data is being collected by the target \\  \hline
             \textcolor{blue}{Medium} & How the data will be collected by the target \\  \toprule
              
        \end{tabular}}
    }
    \hspace{1em}
    \subfloat[\label{fig:annotation-relations} Relation between entities]{\includegraphics[width=0.4\textwidth]{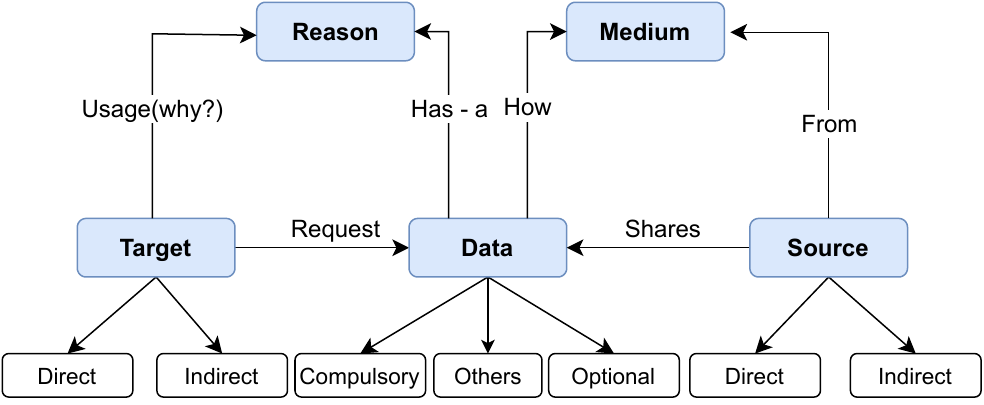}}
    \\
    \vspace{1em}
    \subfloat[\label{fig:annotation-example} Example of an annotated paragraph with labelled entities.]{
    \resizebox{\textwidth}{!}{
    \doublespacing
    
    \begin{tabular}{p{50em}}
    \toprule 
        \textbf{Paragraph:} When \textcolor{red}{\{{\em \ul{you}}\}$_\text{source-direct}$} \textcolor{blue}{\{{\em \ul{visit the site}}\}$_\text{medium}$}, \textcolor{teal}{\{{\em \ul{we}}\}$_\text{target-direct}$} also collect \textcolor{purple}{\{{\em \ul{web site usage information , the type and version of browser and operating system}}\}$_\text{data-compulsory}$} \textcolor{red}{\{{\em \ul{you}}\}$_\text{source-direct}$} use, \textcolor{purple}{\{{\em \ul{if you arrived at trainchinese.com via a link from another website , the URL of the linking page}}\}$_\text{data-compulsory}$}. \textcolor{teal}{\{{\em \ul{We}}\}$_\text{target-direct}$} use this information to \textcolor{violet}{\{{\em \ul{ensure our site is compatible with the browsers used by most of our visitors}}\}$_\text{reason}$} and to \textcolor{violet}{\{{\em \ul{improve the customer experience}}\}$_\text{reason}$}. \\ \toprule
    \end{tabular}}
    }
\caption{Annotation labels with their definitions, inter-relationships, and an annotated instance. Best viewed in colors.}
\label{fig:labels-details}
\end{figure*}

    


    

\paragraph{Motivation and Problem Definition:}
Privacy policies are essential for both businesses and individuals. For businesses, having a privacy policy can protect them from legal issues related to data privacy and usage. On the other hand, it provides transparency to individuals about how their personal information will be managed and protected by the organizations; thus enabling them to make informed decisions prior to registering for the service. Despite its importance, very few users read these lengthy and non-trivial documents and fall prey to their inadvertent consent.

Summarizing these documents is a straightforward remedy of the lengthy document but it needs to ensure that every aspect of the data usage/management must also be present in the summary to make it useful. However, given the complicated nature of the policy document, it's non-trivial to obtain every critical privacy-related information in a summary. For instance, some policy documents define different data items (\textit{viz.} name, age, contact details, etc.) at the beginning of the document but refrain from reporting the management of data items until the end of the document or in different paragraph or context; thus making it challenging for any summarization system to deal with such cases. In such cases, controlled abstractive summarization techniques  \cite{He2020CTRLsumTG,Liu2021ControllableND,zhang2023macsum} can potentially enhance accessibility and transparency by generating concise and coherent summaries of policy documents. 
Acknowledging the severity of the problem, in this paper, we propose an \textbf{Entity-dRiven cOntrlled policy document Summarization} system \textit{aka.} \textbf{\modelname}. \modelname\ operates in two stages: 1) it extracts various entities or data items and their rationales through a BERT and XLNet-based entity-extraction module; and 2) leveraging the extracted entities, it mandates a BART-based summarization module to include these entities and their rationales through a proximal policy optimization (PPO) framework.    

We developed a dataset, namely \dataset, of 1900 policy documents and manually annotated them with abstractive summaries along with privacy-related entities and their rationales. At first, we mark all entities present in the document and also identify what, why, and how they are being collected. To achieve this, we proposed and followed a schema for the identification of critical privacy-related information in a policy document as depicted in Figure~\ref{fig:annotation-relations} (c.f. Section \ref{sec:dataset} for details). It includes the \textbf{\em data} being collected, who would be the \textbf{\em source} of data, through which \textbf{\em medium} data will be collected, who will consume (\textbf{\em target}) the data, and what is the \textbf{\em reason} of data collection. An example with annotated entities is shown in Figure \ref{fig:annotation-example}. In the next step, we write a summary of the document mandating the presence of the entities and their rationales along with other relevant information.

Our experimental results demonstrate that \modelname\ achieves state-of-the-art performance on the proposed \dataset\ dataset against several baseline systems. We also perform qualitative and error analyses to assess the capability of \modelname\ in ensuring various aspects of the privacy-related information in the generated summaries.

\paragraph{Contribution:} The main contributions of this paper are summarized as follows:
\begin{itemize}[noitemsep, nolistsep, topsep=0pt, leftmargin=1em]
  \item We propose a BART-based entity-driven controlled policy document summarization (\modelname) to mitigate the concerns of general public over the data privacy and security issues. 
  \item To identify privacy-related relevant information in a policy document, we developed an entity extraction model, \textbf{E}ntity \textbf{E}xtraction from \textbf{P}olicy \textbf{D}ocuments (\nermodelname).
  \item We introduce a personalized loss function and a reinforcement learning framework using Proximal Policy Optimization (PPO) to manage the relevance and length of the generated summaries. 
  \item We introduce a new dataset (\dataset) of privacy policy documents with their summaries and privacy-bounded entities and rationales. 
  \item We also establish performance benchmarks for the proposed approach against several baselines.
  \item Finally, we perform qualitative and error analyses to assess the quality of summaries.
\end{itemize}

\paragraph{Reproducibility:} Code and dataset are available at \url{https://github.com/joykirat18/EROS}.

\section{Related Work}
Pretrained encoders have become pivotal in recent summarization approaches. \citet{liu2020presumm} introduced a BERT-based unsupervised text summarization model, achieving state-of-the-art performance on benchmark datasets. \citet{wang2021fine} explored fine-tuning strategies for language models like BERT and GPT, revealing substantial performance gains through limited labelled data utilization. \citet{dong2021hierarchical} proposed a hierarchical transformer model for summarizing lengthy documents, achieving leading results on multiple benchmarks. \citet{zhang2020pegasus} devised PEGASUS, leveraging gap sentence extraction and transformer-based gap filling pre-training, attaining state-of-the-art performance on various benchmarks. These studies depict the impactful role of pre-trained encoders in advancing summarization techniques. 

Entity extraction is a fundamental task in information extraction. The problem has been modelled in multiple ways such as sequence labelling \citet{ DBLP:journals/corr/abs-2004-07493, bui2021automated}, span level prediction \citet{eberts_ulges_2020, zhong2020frustratingly, zhu-li-2022-boundary}, question answering \citet{li-etal-2020-unified} as well as dependency parsing task \citet{yu-etal-2020-named}.

Recently, a paradigm shift has been observed from sequence labelling tasks to span-based prediction of entities. In span-based task, such as \citet{eberts_ulges_2020}, all possible spans are selected and further classified whether that span represents an entity or not followed by relation classification, if required. To tackle the problem of exact spans being treated correctly and partial spans being treated incorrectly, \citet{zhu-li-2022-boundary}, proposes a way to regularize span-based prediction tasks. The annotated spans are assigned a full probability and the nearby tokens are also assigned some probability of being correct. \citet{zhong2020frustratingly} extracts entities along with relation instead of the traditional approach of extracting entities and then using the extracted entities for relation classification.   

Reinforcement learning (RL) has gained traction in summarization \cite{wang2018hierarchical, wan2018improving}. \citet{rondeau2018reinforcement} introduced RL-driven translation with simulated human feedback. \citet{liu2020learning} addressed RL's reward scarcity using human feedback. \citet{gunasekara2021using} presented a versatile framework using RL for abstractive summarization through question-answering rewards. These efforts highlight RL's effectiveness in improving summarization. In another work, \citet{He2020CTRLsumTG} proposed  controlled summarization to let the user interact with the summary either in the form of keywords or direct prompts. \citet{saito2020abstractive} develops a combination model consisting of a saliency model that extracts a token sequence from a source text and a seq-to-seq model that takes the sequence as an additional input text.

In comparison to existing works, our approach integrates pre-trained encoders, reinforcement learning, and modified loss functions. Our model incorporates a penalty mechanism in addition to a refined BART architecture for controlled summarization. In order to provide more exact and controlled summaries, this combined technique builds on the advantages of each component resulting in more precise and controlled summary generation. Additionally, our method uniquely incorporates insights from an Entity Extraction task, enhancing the model's ability to capture information from policy documents.

\section{Dataset Construction} \label{sec:dataset}
To the best of our knowledge, the domain of privacy-driven policy document summarization has not been studied so far; hence, we recognize the need for a dataset to facilitate our research and thereby, develop \dataset, tailor-made dataset for the policy document summarization. 

\paragraph{Data collection and Filtering:} We collect policy documents of different websites curated by \citet{amos2021privacy}. These documents outline how websites manage (i.e., collect, use, or disclose) personal information. After collection, we observed several issues and hence, applied a filter to discard policy documents as follows:  
\begin{itemize}[noitemsep, nolistsep, topsep=0pt, leftmargin=*]
  \item Many websites have identical policy documents, we discard all but one.
  \item In case there are URLs linking to other websites, disregard them. 
  \item Skip documents that lack meaningful/significant information or are incomplete.
  \item Refrain from including any policy content that is not relevant to the topic at hand.
\end{itemize}

\noindent Subsequent to the filtering process, 1921 policy document remains in \dataset.

\paragraph{Data Annotation:} To facilitate the entity-driven controlled summarization, we need two sets of annotations: a) identification of privacy-related entities; and b) a summary of the document. Considering the users' concerns, we identified five fundamental entities regarding data privacy and security and proposed a schema (c.f. Figure \ref{fig:annotation-relations}) to capture their relationships:
\begin{itemize}[noitemsep, topsep=0pt, leftmargin=*]
    \item \textbf{Data:} It defines the type of information that an organization usually collects -- \textit{name}, \textit{email}, \textit{contact number}, \textit{address}, \textit{location}, \textit{photos}, \textit{system details}, \textit{browsing history}, \textit{search queries/patterns}, \textit{keystrokes}, etc. Further, we observe that some of these data are compulsory as part of the service agreement, while others are optional and users can deny access without any interruption in service. We identify data items according to these two categories, i.e., `\textit{data compulsory}' and `\textit{data optional}'. Further, we identify data items as `\textit{data others}' which are not associated with a target requesting the data explicitly may not highlight its usage, e.g., the data item `\textit{personal information}' in the sentence `\textit{We may share personal information with our clients}'.
    \item \textbf{Source:} It signifies the provider of the information. While a majority of the time, the user (e.g., `\textit{you}') is the direct source, in some cases, the source can be indirect, e.g., "inviting your friends to join the website by sharing contact information "(\textit{friends} will be source indirect), "requiring someone else information to ship products to their address"(\textit{someone else} is source indirect). These entities are very low in number and are majorly associated with sharing someone else information.
    \item \textbf{Medium:} It defines the way data is collected such as `\textit{while visiting the website}', `\textit{responding to a survey}', `\textit{filling a form}', etc. 
    \item \textbf{Target:} It specifies who will consume the data. Similar to the source entity, a target can be direct (\textit{the organization itself}) or indirect (\textit{any third-party vendor outside the organization}). Though the direct target is somewhat benign as the users know their data are being used for some specific purposes by the service provider, the indirect target can be extremely detrimental as there is no transparency about the usage of data in an unknown capacity. 
    \item \textbf{Reason:} It clarifies the purpose of data collection by the parent organization such as `\textit{improving customer experience}'. We observe that with indirect targets, reasons are usually hidden or extremely vague.
\end{itemize}

\noindent Following the above schema, two annotators\footnote{Annotators were undergraduate student volunteers and in the age group of 20-30.} with good English proficiency annotated the whole dataset using LabelBox\footnote{\url{https://labelbox.com/}} as the annotation tool. At first, we tokenize the sentences using NLTK tokenizer \citet{bird2009natural}, and subsequently, for each identified entity, we record their start and end indices as span. Further, to ensure the consistency of annotations between them, the annotators independently annotated a small set of documents in the pilot phase and discussed their common understanding. Next, they annotated $10$ documents separately and achieved a Cohen Kappa inter-annotator agreement score of $0.74$. 
Subsequently, we manually annotate $150$ documents having $8000$ sentences with $9094$ distinct entity labels. In the next step, we learn an entity-recognition model trained on the annotated $150$ documents to obtain pseudo entity labels for the remaining $1771$ documents. An illustrative example of annotation for a paragraph of the document is presented in Figure \ref{fig:annotation-example}. Table \ref{tab:entity-freq} lists the distribution of the entities in \dataset.


 

\begin{table}[t]
    \centering
    \resizebox{0.7\columnwidth}{!}{
\begin{tabular}{llcc} 
\toprule
 & & \textbf{Train} & \textbf{Test} \\ \midrule
 
 \multirow{12}{*}{\rotatebox{90}{\textbf{Documents}}} & \#Doc & 1536 &385 \\  
 & \#Sent in Doc & 116251 & 27487\\ 
 & Avg. token/Doc & 1455.6 & 1375.82\\
 & Total entities (Doc) & 130856 &31146\\ 
 & \quad -- Data other &  26836&6458\\
 & \quad -- Data Compulsory &  7715&1866\\
 & \quad -- Data Optional & 274 & 68 \\
 & \quad -- Reason &  10720&2451\\
 & \quad -- Medium &  11924&2922\\
 & \quad -- Target Direct &  33927&8159\\
 & \quad -- Target Indirect &  12421&3129\\
 & \quad -- Source Direct &  27002&6080\\
 & \quad -- Source Indirect & 37 & 13  \\  \midrule
 
 
 \multirow{12}{*}{\rotatebox{90}{\textbf{Summary}}}  
 & \#Sent in Summ & 18691 &4500\\ 
 & Avg. token/Summ & 198.9 & 192.46 \\  
 & Total entities (Summ) & 28197 &6952 \\ 
 & \quad -- Data other & 6244&1580\\
 & \quad -- Data Compulsory & 1784&497\\
 & \quad -- Data Optional & 0 & 0 \\
 & \quad -- Reason & 1449&338\\
 & \quad -- Medium & 2469&586\\
 & \quad -- Target Direct & 7991&1938\\
 & \quad -- Target Indirect & 2453 &648\\
 & \quad -- Source Direct & 5807 &1365 \\
 & \quad -- Source Indirect & 0 & 0  \\ \bottomrule

\end{tabular}}
    \caption{Dataset statistics of \dataset.}
    \label{tab:entity-freq}
\end{table}

In the second stage, we focus on manually annotating $1921$ documents with a brief yet informative summary. We provide clear guidelines to our annotators, encouraging them to extract key data-related phrases and concepts from the document while preserving the essence of the content in a concise and clear manner. The goal is to keep the summaries as crisp and to the point as possible. We emphasize using simple and everyday language, steering away from complex or overly technical terminology. Our aim is to make the summaries accessible and easily understandable to a broad audience. This approach ensures that the information conveyed in the summaries remains approachable and digestible.
In terms of document length, we've found that policy documents typically contain an average of $1500$ tokens. On the other hand, the annotated summaries, following our guidelines, average about $200$ tokens. This significant reduction in token count maintains a balance between providing necessary information and keeping the summaries succinct and user-friendly.
\begin{figure*}[h!]
  \centering
  \includegraphics[width=\textwidth]{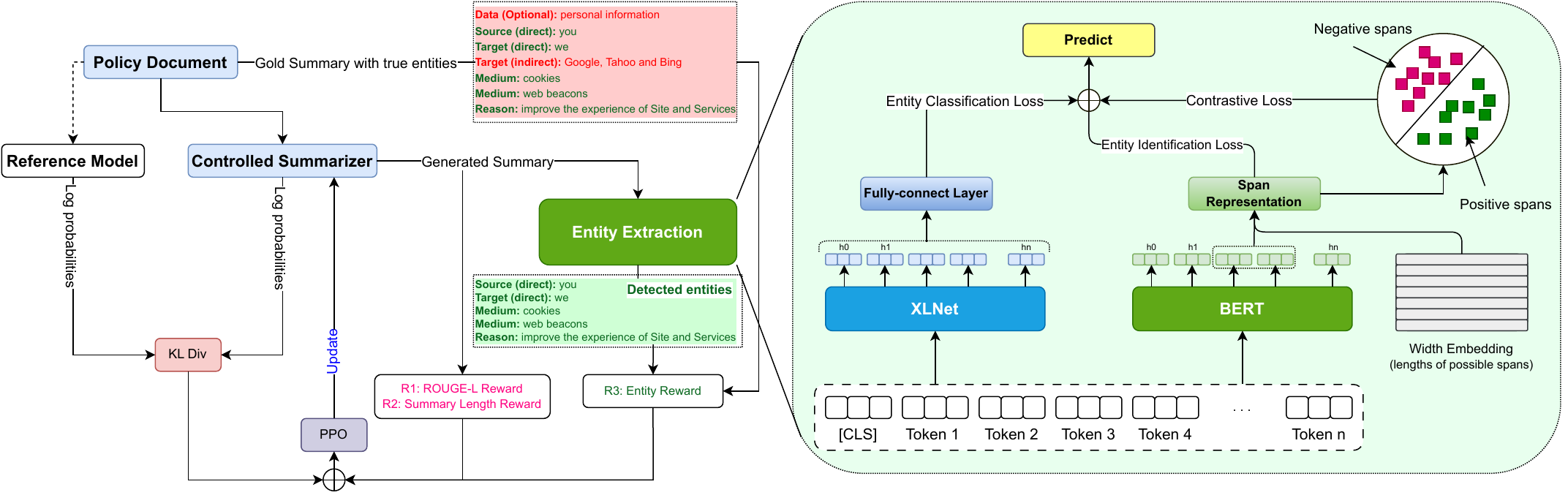}
  \caption{\textbf{Left:} Proposed Model for \modelname. The reference model is a frozen pre-trained BART-based model with modified loss. We initialize the controlled summarization model in a similar way, which is subsequently updated through a PPO framework on a combination of rewards and KL-divergence loss. \textbf{Right:} Entity extraction model jointly learns a entity classification and entity identification module with the assistance of contrastive loss. Further to minimize the effect of false positives in identification, we supplement it with a entity classification module in a joint framework.}
  \label{fig:nerModel}
  \vspace{-3mm}
\end{figure*}

\section{Proposed Methodology - \modelname}
Our proposed controlled summarization model, \modelname, is depicted in Figure \ref{fig:nerModel}. It works in two stages. At first, we train an entity extraction model that aims to predicts all spans of entities in a given document. This model employ BERT-based span prediction framework with the contrastive loss. Additionally, we supplement the prediction via an entity classification model in a joint learning setup. In the second stage, we employ a BART-based summarizer model to generate a summary. To assess the quality of generated summary and to ensure induction of critical entity information in the summary, we introduce our pretrained entity extraction module in the pipeline. The extracted entities are then compared with the true (gold) entities of the reference summary and an entity reward is computed. Moreover, we compute two other rewards (ROUGE-L and summary length) to ensure a relevant and concise summary. The accumulated reward is then added with the KL-divergence score between the log-probabilities of the controlled summarizer and a reference model (a BART-based summarizer trained on a huge out-of-domain corpus). Subsequently, we employ PPO to update the controlled summarizer. This process repeats for a few step till (near-)convergence. In the following subsections, we elaborate on these steps.


\subsubsection*{Entity extraction module}
Recent years have seen a paradigm shift in the task of entity recognition from token-level tagging --which conceptualizes it as a sequence labelling task-- to span-level prediction \citet{fu-etal-2021-spanner}. 

A span in a sentence is represented by the start and end token of a sentence. Given a sentence $X=\left\{x_1, \cdots, x_n\right\}$ with $n$ tokens, we define a span of an entity as $s_i^y=\left\{x_{b_i}, x_{b_i+1}, \cdots, x_{e_i}\right\}$, where $b_i$ and $e_i$ denote the start and end index of the span $s_i$ with a corresponding tag $y \in \{$\textit{source-direct, source-indirect, data-optional, data-compulsory, medium, target-direct, target-indirect, reason}$\}$. 

Spans come in different lengths. To avoid over-fitting for a particular length, we adopt an enumeration strategy, where all the possible $m$ spans with a maximum length $l$ are being considered as valid spans for predicting entities. For example, in the sentence, ``\textit{we will collect name}'' with maximum span length as 4, possible spans are: $s_1^y = \{x_1, x_1\}$, $s_2^y = \{x_1, x_2\}$, $s_3^y = \{x_1, x_3\}$, $s_4^y = \{x_1, x_4\}$, $s_5^y = \{x_2, x_2\}$, $s_6^y = \{x_2, x_3\}$, $s_7^y = \{x_2, x_4\}$, $s_8^y = \{x_3, x_3\}$, $s_9^y = \{x_3, x_4\}$, and $s_{10}^y = \{x_4, x_4\}$. From gold labels, we know that out of these 10 spans, only $s_1$ and $s_4$ are valid spans; therefore, their $y$ labels will be source-target and data-compulsory, respectively. For all other spans, the $y$ labels will be ``invalid (0)''.

We obtain token representation from BERT and subsequently, compute span embeddings as $\mathbf{z}_i^b=\left[\mathbf{h}_{b_i}; \mathbf{h}_{e_i}\right]$
Additionally, to provide information regarding the width of each span, we induce a learnable width encoding vector $z_i^w$ according to their width. Thus, the final representation becomes  $\mathbf{s}_i=\left[\mathbf{z}_i^b ; \mathbf{z}_i^w\right]$.
Our initial experiments showed encouraging results; however, we also observe a significant number of false-positive, especially, for a sentence with no entity at all, in our predictions. To mitigate such issue, we incorporate a binary classification task to identify if the sentence contains an entity in a joint learning framework. 

\paragraph{Contrastive Loss: }
We compute a similarity score between pairs of input spans, and then minimize the distance between similar pairs while maximizing the distance between dissimilar pairs using the contrastive loss:
\begin{align}
\mathrm{P}\left(\mathbf{y} \mid s_i\right)=\frac{\operatorname{score}\left(\mathbf{s}_i, \mathbf{y}\right)}{\sum_{\mathbf{y}^{\prime} \in \mathcal{Y}} \operatorname{score}\left(\mathbf{s}_i, \mathbf{y}^{\prime}\right)}
\end{align}
where $\operatorname{score}\left(\mathbf{s}_i, \mathbf{y}_k\right)=\exp \left(\mathbf{s}_i^T \mathbf{y}_k\right)$, is a function that measures the compatibility between a learnable label representation of the class k $\mathbf{y}_k$ and span $\mathbf{s}_i$.



\paragraph{Span Prediction: }
Finally, the span representations $s_i$ are fed into a softmax function to get the probability considering the label $y$. For optimization, we combine the three losses -- entity identification cross-entropy loss ($\ell_1$), binary classification loss ($\ell_2$), and the contrastive loss ($\ell_3$) through the following weighting mechanism, where $\alpha_1$, $\alpha_2$, and $\alpha_3$ are hyperparameters. 
\begin{align}
\mathcal{L}^{e} = \alpha_1 \ell_1 + \alpha_2 \ell_2 + \alpha_3 \ell_3, \quad \text{where} \sum \alpha_i  = 1
\end{align}


\subsubsection*{Entity-Driven Controlled Summarization} 
\label{Summ}
As the foundational model, we employ BART in our experiment. Further, we induced a modified loss function specially designed for the entity-driven controlled summary generation.  
To elaborate, we obtain gold entities ($e_i$) from the \dataset\ dataset and integrate them into the loss function of the BART model. This entails augmenting the traditional cross-entropy loss with a penalty component derived from the extracted entities. It enables the BART model to comprehend the presence and importance of entities in the summaries, thereby refining its summary generation capabilities while maintaining control over the process. Mathematically it can be seen as follows:
\begin{align}
  \label{eq:cel}  
  \text{CE} & = - \sum(y \cdot \log(x)) \nonumber \\
  TP & = \sum_{e_i} (1.0 - \text{step}(e_i \in S_G)) \nonumber \\
  \mathcal{L}^{s} & = \lambda \cdot \text{CE} + (1-\lambda) \cdot \text{TP} 
\end{align}
where CE, TP, $\lambda$, and $S_G$ are the cross-entropy, token penalty score, weight of the loss, and the generated summary, respectively. We compute TP by penalizing the model for each missing entity $e_i$ in $S_G$. The step function will return 1 only if the entity is part of the summary, otherwise, a value of 0 will be returned. 


Further to supplement the controlled summary generation process, we adopted a feedback mechanism, in the form of reinforcement learning, to reward/penalize the model for inducing/not-inducing the privacy-related entities in the summaries. We use proximal policy optimization (PPO) to enforce the model to improve the generation quality. 
First introduced by \citet{DBLP:journals/corr/SchulmanWDRK17}, PPO refines policy adjustments by combining ratio-based enhancement with a clipped surrogate objective; thus, ensuring controlled updates. Incorporating an auxiliary value function, PPO enhances policy updates by estimating advantages and rewards more accurately, particularly in complex scenarios.

The proposed model shown in Figure \ref{fig:nerModel}, contains a policy model (i.e., the controlled summarizer model that is being trained), a reference model, a reward model, and a value function. The value function is used to describe the reward at timestep $t$. On the other hand, the reference model is used to 
calculate the KL divergence between the original model and the policy model. The main idea is to ensure that the active model does not deviate a lot from its original distribution. 

\paragraph{Reward calculation:} We compute three rewards to maintain the coverage, conciseness, and relevance in the summary. The coverage reward ensures the readability of the generated summary --we compute the ROUGE-L score, which is based on the longest common subsequence (LCS) between two sequences, as the first reward ($R1 = \text{ROUGE-L}(S_G, S_R)$). A longer LCS indicates that the generated summary conveys similar meaning and concepts as the reference summary. The conciseness reward ($R2$) limits the model to generate an adequate length summary and avoid generating lengthy jargon. 
\begin{align} 
R_2 = \frac{1 - |(len(S_G) - len(S_R))|}{max(len(S_G), len(S_R))}
\end{align}
Finally, we compute the entity reward ($R3$) as follows: Let \(E_{total}\) be the total number of entities predicted from the generated summary, and \(E_{correct}\) and \(E_{incorrect}\) be the number of entities present and not present in the gold summary respectively.  
\begin{align} 
R_3 = \frac{E_{correct} - \beta * E_{incorrect}}{E_{total}}
\end{align}
where $\beta$ is a negative factor for penalizing incorrect entities. We set $\beta=0.3$ for our experiments.

\begin{table}
    \centering
    \resizebox{\columnwidth}{!}{
    \begin{tabular}{l:l|l:l|l:l}
        \toprule
        \multicolumn{2}{c|}{\textbf{Training \textit{\nermodelname}}} & \multicolumn{2}{c|}{\textbf{Training \textit{\modelname}}} & \multicolumn{2}{c}{\textbf{Generating Text (\textit{\modelname})}} \\
        \textbf{Hyper-parameter} & \textbf{Value} & \textbf{Hyper-parameter} & \textbf{Value} & \textbf{Hyper-parameter} & \textbf{Value} \\
        \midrule
        \textit{n\_class} & 10 & \textit{warmup} & 0.1 & \textit{max\_seq\_length} & 1024 \\
        \textit{bert\_dropout} & 0.2 & \textit{learning\_rate} & 5.41e-6 & \textit{min\_new\_tokens} & 200 \\
        \textit{xlnet\_dropout} & 0.2 & \textit{adaptive\_kl\_coef} & True & \textit{top\_p} & 0.9 \\
        \textit{LR} & 1e-5 & \textit{gamma} & 0.99 & \textit{do\_sample} & True \\
        \textit{maxlen} & 512 & \textit{batch\_size} & 8 & \textit{top\_k} & 10 \\
        \textit{maxnorm} & 1.0 & \textit{mini\_batch\_size} & 2 & \textit{use\_cache} & True \\
        \textit{batchSize} & 4 & & & \textit{num\_beams} & 1 \\
        \textit{max\_spanLen} & 10 & & & & \\
        \textit{spanLen\_emb\_dim} & 300 & & & & \\
        \(\alpha_1\) & 0.5 & & & & \\
        \(\alpha_2\) & 0.25 & & & & \\
        \(\alpha_3\) & 0.25 & & & & \\
        \textit{l} & 10 & & & & \\
        \bottomrule
    \end{tabular}}
    \caption{Hyperparameters for training \textit{\nermodelname}, \textit{\modelname}, and generating summary from \textit{\modelname}}
    \label{tab:hyperparameter}
\end{table}


\begin{table*}[!ht]
    \centering
    \resizebox{0.8\textwidth}{!}{
    \begin{tabular}{lcccccccccc}
        \toprule
        \multirow{2}{*}{\textbf{Model}} &
        \multicolumn{3}{c}{\textbf{Rouge}} & & \multicolumn{4}{c}{\textbf{BLEU}} & \multirow{2}{*}{\textbf{METEOR}} & \multirow{2}{*}{\textbf{BS}} \\ \cmidrule{2-4} \cmidrule{6-9}
         & \textbf{R1} & \textbf{R2} & \textbf{RL} & & \textbf{B1} & \textbf{B2} & \textbf{B3} & \textbf{B4} & \textbf{} \\
        \midrule
        Extractive Oracle \citet{} & 0.43 & 0.30 & 0.42 & & 0.14 & 0.12 & 0.11 & 0.10 & 0.25 & 0.881\\
        Bert2Bert \citet{} & 0.25 & 0.04 & 0.22 & & 0.22 & 0.08 & 0.10 & 0.16 & 0.25 & 0.818\\
        T5-Summarizer \citet{} & 0.44 & 0.24	& 0.42 & & 0.32 & 0.24 & 0.19 & 0.17 & 0.34 & 0.877 \\
        PEGASUS \citet{} & 0.35 & 0.17 & 0.32 & & 0.18 & 0.12 & 0.09 & 0.08 & 0.23 & 0.859 \\
        BART \citet{} & 0.46 & 0.29 & 0.44 & & 0.31 & 0.25 & 0.22 & 0.20 & 0.33 & 0.882 \\       
        CTRL-SUM [bigpatent]  & 0.200 & 0.037 & 0.180 & & 0.174 &  0.067 & 0.022 & 0.008 & 0.142 & 0.798 \\
        CTRL-SUM [cnndm] & 0.340 & 0.148 & 0.318 & & 0.266 & 0.164 & 0.121 & 0.099 & 0.246 & 0.837 \\
        CTRL-SUM [\dataset] & 0.355 & 0.124 & 0.327 & & 0.244 &  0.134 & 0.077 & 0.044 & 0.216 & 0.848 \\ \midrule
        T5-Loss & 0.45 & 0.26 & 0.43 & & 0.32 & 0.24 & 0.19 & 0.17 & 0.34 & 0.877 \\
        PEGASUS-Loss & 0.38 & 0.21 & 0.36 & & 0.23 & 0.17 & 0.14 & 0.12 & 0.26 & 0.865 \\
        BART-Loss & 0.48 & 0.31 & 0.46 & & 0.31 & 0.25 & 0.22 & 0.20 & 0.34 & 0.889 \\ \midrule
        \modelname & \bf 0.519 & \bf 0.341 & \bf  0.500 & & \bf 0.424 & \bf 0.337 & \bf 0.292 & \bf 0.262 & \bf 0.438 & \bf 0.891\\ 
        \hdashline
        \modelname\ w/o R1 & 0.434 & 0.232 & 0.412 & & 0.351 & 0.241 & 0.185 & 0.148 & 0.33 & 0.871\\
        \modelname\ w/o R2 & 0.414 & 0.211 & 0.395 & & 0.328 & 0.217 & 0.16 & 0.124 & 0.295 & 0.8648\\
        \modelname\ w/o R3 & 0.510 & 0.324 & 0.491 & & 0.412 & 0.322 & 0.275 & 0.244 & 0.415 & 0.888\\
        \bottomrule
    \end{tabular}}
    \caption{Rouge, Bleu, Meteor and BertScore scores for baselines and our proposed \modelname{} model. The term ``Loss" signifies application of a customized loss function (c.f. equation \ref{eq:cel}). CTRL-SUM [$D$] signifies that the model is trained on dataset $D$.}
    \label{tab:baseline_results}
\end{table*}

\section{Experiments and Results}
\label{sec:Experiements}
We train \modelname\ on 1536 documents, while we use 385 documents for evaluating the performance. A summary of the hyperparameters used for training is listed in Table \ref{tab:hyperparameter}. We evaluate both components of our model separately and their results are furnished in Tables \ref{tab:baseline_results} and \ref{tab:nerResults}, respectively -- we compute precision, recall, and F1-score for the entity extraction module, we measure the quality of summarization through BLEU, ROUGE, METEOR, and BERTScore. We also perform extensive comparative analysis against the following baselines for both models. In all cases, we re-train/fine-tune these models on \dataset.


\paragraph{Baselines:}
\begin{itemize}[ topsep=0pt, leftmargin=1em]
\setlength{\itemsep}{0pt}    
    \item \textbf{Entity Extraction:} We evaluate \nermodelname\ against six entity extraction models covering both sequence-labelling and span-based frameworks:
    \textbf{Sequence-labelling: } Taggin each token as \textbf{B}egin, \textbf{I}ntermedite, or \textbf{O}ther -- 
    \textbf{BERT} \citet{devlin2018bert},  
    \textbf{SpanBERT} \citet{joshi2020spanbert}, and 
    \textbf{PrivBERT} \citet{srinath-etal-2021-privacy}. 
    \textbf{Span-based:} Start/end index-based extraction models -- \textbf{Boundary Smoothing} \citet{zhu-li-2022-boundary}, 
    \textbf{PURE} \citet{zhong2020frustratingly}, 
    \textbf{SPERT} \citet{eberts_ulges_2020}. 
    \item \textbf{Entity-Driven Controlled Summarization:} For the controlled summarization model, we compare \modelname's effectiveness against the following baseline approaches:
        \textbf{Extractive Oracle} \cite{hirao-etal-2017-enumeration}: It employs extractive methods to gather essential information from the source text for summarization. This model offers efficiency by avoiding new sentence generation and at the expense of missing overall context and flow.
        \textbf{PEGASUS} \cite{zhang2020pegasus}: Pre-trained transformer-based sequence-to-sequence architecture designed by Google AI. The model uses a novel pre-training objective known as "gap-sentences generation". 
        \textbf{CTRL-SUM} \cite{He2020CTRLsumTG}: It uses control tokens at inference time to enable user to control the summary. In the original work, it has been trained on two huge datasets -- BIGPATENT patent documents \cite{sharma2019bigpatent} and CNN/Dailymail news articles \cite{hermann2015teaching}. Further, in this work, we also train CTRL-SUM on \dataset. In addition, we also compare with \textbf{Bert2Bert} \cite{chen-etal-2022-bert2bert}, \textbf{T5} \cite{raffel2020exploring}, and \textbf{BART} \cite{lewis2019bart} summarizers. 
\end{itemize}

\begin{table}[t]
\centering
\resizebox{\columnwidth}{!}{
\begin{tabular}{llccc } 
 \toprule
 & & \bf Precision & \bf Recall & \bf F1-score \\ 
 \midrule
 \multirow{3}{*}{\rotatebox{90}{\textbf{BIO}}} & BERT & 0.31 & 0.38 & 0.34 \\ 
 & SpanBERT & 0.31 & 0.39 & 0.35 \\ 
 & PrivBERT & 0.37 & 0.44 & 0.40 \\ 
 \midrule
 \multirow{7}{*}{\rotatebox{90}{\textbf{Span Based}}} & SPERT & 0.10 & \textbf{0.68} & 0.17 \\ 
 & Boundary Smoothing & 0.40 & 0.48 & 0.44 \\ 
 & PURE & 0.35 & 0.12 & 0.17 \\
 \cmidrule{2-5}
 & SpanNER & 0.49 & 0.56 & 0.52 \\
 & SpanNER + Identification & 0.47 & 0.66 & 0.55 \\
 \cmidrule{2-5}
 & \textbf{\nermodelname} & \textbf{0.54} & 0.62 & \textbf{0.58} \\
 & \quad -- Identification & 0.48 & 0.57 & 0.52 \\
 \bottomrule
 
\end{tabular}}
    \caption{Comparative results of entity extraction module.
}
    \label{tab:nerResults}
\end{table}

\paragraph{Result Analysis:} Table \ref{tab:nerResults} contains the comparative result of \nermodelname\ and various baselines.
PrivBert reports the best F1-score of $0.40$ in the sequence-labelling framework (i.e., in a BIO setup), whereas, BoundarySmoothing yields $+4\%$ better F1-score at $0.44$ in the span-based setting. Further, SpanNER, with the identification module,  records the best F1-score of $0.58$ among all baselines. In comparison, \nermodelname\ reports the state-of-the-art performance at $0.58$ F1-score -- an increment of $+3\%$ over the best baseline. We also observe the effect of the entity identification module on the overall performance -- a decrement of $-6\%$ is observed on removing the identification component from \nermodelname. 

\begin{table*}
\centering
\resizebox{\textwidth}{!}{
\begin{tabular}{p{70em}}
\toprule
\textbf{Ref Summary: } When \textcolor{red}{you} \textcolor{blue}{register with the Site or use any of our Services}, you may be asked to provide \textcolor{teal}{us} with \textcolor{brown}{Personal Information}. It is entirely optional to provide this information. If you do not provide the requested information, you may be unable to use some or all of the Site's features. To \textcolor{violet}{improve the Site and Services}, we use \textcolor{blue}{cookies} (a small text file placed on your computer to identify your computer and browser) and \textcolor{blue}{Web beacons} (electronic files placed on a Web site that monitor usage). If you delete or disable cookies, some features of the Site or Services may not function properly. \textcolor{red}{You} have the option of making some of your \textcolor{brown}{Personal Information} available to others. \textcolor{orange}{Users of the Site} and commercial search engines such as \textcolor{orange}{Google}, \textcolor{orange}{Yahoo!}, and \textcolor{orange}{Bing} may access the information you choose to make available. Some \textcolor{orange}{third-party services} may provide \textcolor{teal}{us} with information from your accounts, allowing us to \textcolor{violet}{improve and personalise your experience on the Site}. When \textcolor{red}{you} \textcolor{blue}{visit our Site}, we may allow \textcolor{orange}{third-party companies} to serve ads and/or collect certain \textcolor{brown}{anonymous information}.... \\ \hdashline 

\textbf{Entities in Ref summary:} \textcolor{purple}{Data compulsory}: [e-mail, personally identifiable information]; \textcolor{brown}{Data others}: [personal information]; \textcolor{red}{Source Direct}: [you, info@day-finder.com]; \textcolor{teal}{Target Direct}: [we, us]; \textcolor{orange}{Target Indirect}: [service providers, third parties]; \textcolor{blue}{Medium}: [register with the Site or use any of our Services, cookies, web beacons, visit our site]; \textcolor{violet}{Reason}: [improve the site and services, improve and personalise your experience on the Site, Usage Data regarding the Site and Services]. \\ \midrule

\textbf{Entities in BART summary:} \textcolor{purple}{Data compulsory}: [None]; \textcolor{brown}{Data others}: [None]; \textcolor{red}{Source Direct}: [you, info@day-finder.com]; \textcolor{teal}{Target Direct}: [we, us]; \textcolor{orange}{Target Indirect}: [None]; \textcolor{blue}{Medium}: [using the site or services, cookies, web beacons, visit our site]; \textcolor{violet}{Reason}: [improve the experience of the site and services].\\ \midrule

\textbf{Entities in BART-Loss summary:} \textcolor{purple}{Data compulsory}: [e-mail, personally identifiable information]; \textcolor{brown}{Data others}: [None]; \textcolor{red}{Source Direct}: [you]; \textcolor{teal}{Target Direct}: [we, us]; \textcolor{orange}{Target Indirect}: [service providers]; \textcolor{blue}{Medium}: [cookies, web beacons,]; \textcolor{violet}{Reason}: [None]. \\ \midrule

\textbf{Entities in \modelname\ w/o R3 summary:} \textcolor{purple}{Data compulsory}: [e-mail, personally identifiable information]; \textcolor{brown}{Data others}: [personal information]; \textcolor{red}{Source Direct}: [you]; \textcolor{teal}{Target Direct}: [we, us]; \textcolor{orange}{Target Indirect}: [service providers]; \textcolor{blue}{Medium}: [register with the Site or use any of our Services, cookies, web beacons, visit our site]; \textcolor{violet}{Reason}: [None]. \\ \midrule

\textbf{Entities in \modelname\ summary:} \textcolor{purple}{Data compulsory}: [e-mail, personally identifiable information]; \textcolor{brown}{Data others}: [personal information]; \textcolor{red}{Source Direct}: [you, info@day-finder.com]; \textcolor{teal}{Target Direct}: [we, us]; \textcolor{orange}{Target Indirect}: [service providers, third parties]; \textcolor{blue}{Medium}: [register with the Site or use any of our Services, cookies, web beacons, visit our site]; \textcolor{violet}{Reason}: [Usage Data regarding the Site and Services].\\ \bottomrule
\end{tabular}}
\caption{Qualitative analysis of generated summaries. Due to space constraints, we could not provide the generated summaries of models. 
\textcolor{red}{source direct}, 
\textcolor{teal}{target direct}, 
\textcolor{blue}{medium - blue}, %
\textcolor{violet}{reason}, 
\textcolor{brown}{data}, 
\textcolor{purple}{data compulsory}, 
\textcolor{orange}{target indirect}, 
\textcolor{olive}{data optional}. Best viewed in color. 
}
\label{tab:ann-table}
\vspace{-2mm}
\end{table*}


For the controlled summarization task, we furnish the results in Table \ref{tab:baseline_results}. We compute the traditional ROUGE, METEOR, BLEU and BertScore scores to evaluate the generated summaries. Pretrained Ctrl-sum \citet{He2020CTRLsumTG} failed to perform well when compared to fine-tuned models.  Among all baselines (except with the modified loss, TP (c.f. Section \ref{Summ})), BART reports the best performance across the three metrics -- ROUGE-L score ($0.44$), BLEU-4 ($0.20$), METEOR ($0.33$) and BertScore($0.882$). Further, we observe that the incorporation of modified TP loss obtains comparable results in the majority of the cases and better several setups -- 
ROUGE-L: T5, BART, and PEGASUS improved; BLEU-4: PEGASUS improved; METEOR: BART and PEGASUS improved and BertScore: BART and PEGASUS improved.
On the other hand, \modelname\ yields the best scores across all metrics -- improvement of $+0.04$ in ROUGE-L ($0.50$), $+0.062$ in BLEU-4 ($0.262$), $+0.098$ in METEOR ($0.438$) and $+0.0017$ in BertScore ($0.8907$).

\begin{table}[ht]
    \centering
    {
    \begin{tabular}{lcc}
    \toprule
       \bf Model  & \bf En / Sum & \bf Len dev\\
       \midrule
        Gold Summary & $18.057$ & - \\ \hdashline
        BART & $10.329$ & $-0.2308$\\
        BART-Loss & $10.355$ & $-0.266$\\
        \modelname & \bf $14.436$ & \bf $+0.163$\\
       \bottomrule 
    \end{tabular}}
    \caption{Rate of entities and conciseness of the generated against the gold summary.}
    \label{tab:extraEvaluation}
\end{table}

Table \ref{tab:extraEvaluation} shows a comparative analysis on the conciseness and the rate of captured entities in the generated summaries. We observe $+4\%$ increase in rate of entities in \modelname\ against the two baselines.

\paragraph{Human Evaluation: } We performed a human evaluation on a subset of randomly chosen samples from the \dataset's test set. We compare the summaries of \modelname\ and two baselines i.e., BART and BART-Loss. We ask our evaluators to assess the generated summaries against the reference summaries on four parameters -- the informativeness of the summary (INF), its conciseness (CON), its fluency and grammatical correctness (FL), and the inclusion of relevant entities (EC).  The first two metrics assess the quantity of the information content in the generated summary, whereas the third metric ensures the linguistics quality of the summary. Additionally, the last metric explicitly evaluates the presence of privacy-related entities in the summary. For each parameter, all evaluators assign a rating on a scale of 1 (worst) to 5 (best) based on the quality of the summaries. 
\begin{table}
    \centering
    \begin{tabular}{lcccc}
    \toprule
       \bf Model  & \bf INFO & \bf CON & \bf FLU &\bf EC\\
       \midrule
        BART & 3.0 & 3.35 & 3.75 & 2.90 \\
        BART-Loss & 3.75 & \bf 3.70 & 4.0 & 3.40\\
        \modelname & \bf 4.20 & 3.15 & \bf 4.05 & \bf 4.15\\
       \bottomrule 
    \end{tabular}
    \caption{Human evaluation on \textit{informativeness}, \textit{conciseness}, \textit{fluency}, and \textit{entity coverage}.}
    \label{tab:humanevaluation}
\end{table}
Subsequently, we aggregate the scores through averaging and report the observations in Table \ref{tab:humanevaluation}. We observe that \modelname\ outperforms the other two baseline models in three out of four metrics. It records a comparatively inferior score for conciseness, suggesting that \modelname's summaries are relatively lengthier than others. However, it is better in informativeness, grammatical correctness, and inclusion of relevant entities. 

\section{Conclusion}
In this work, we presented a novel approach for abstractive summarization of privacy policy documents.  Our approach aimed to address the challenge of generating controlled and informative summaries that capture the essence of complex privacy policies. To achieve this, we introduced a customized loss function and incorporated a reinforcement learning framework, enabling us to optimize the relevance of the generated summaries. To facilitate the evaluation and advancement of research in this domain, we also introduced a new dataset for controlled summarization generation. The experimental results obtained from our comprehensive evaluations highlight the effectiveness of the proposed approach. Our model achieved state-of-the-art performance on the custom dataset. The controlled generation of summaries allows for improved accessibility and transparency for users, enabling them to quickly grasp the key points of privacy policies without getting overwhelmed by excessive information. The findings of our work demonstrate the potential of our approach to make a significant impact in the field of privacy policy summarization. By addressing the critical need for concise and user-friendly representations of privacy policies, we contribute to enhancing user understanding. The implications of our work extend to various domains where privacy policies play a crucial role, including data protection, online services, and legal compliance.

\section*{Limitation}
Our research initiative necessitates the utilization of proprietary and sensitive data, encompassing company privacy policies, to train our specialized model. This dataset is comprehensive, containing detailed policies of various companies. This inclusion is vital for our model to effectively understand and learn the intricacies of privacy policies and their implementations.
However, the training process for our model is computationally intensive, requiring substantial computing resources and time. The utilization of a reinforcement learning approach, while effective, significantly extends the duration needed for the training phase. This computational demand poses a notable challenge in terms of time and resource allocation.
We understand that the relevant entities, such as data items, target usage, etc., are extremely critical in comprehending the policy document and hence, any model must include these components in the generated summaries. However, in some cases, the generated summaries may not be exhaustive in terms of these non-trivial entities and care must be taken prior to agreeing to the terms and conditions of the policy document. 

\section*{Ethical Consideration}
In this section, we address the ethical concerns and safeguards associated with our research on privacy policy text summarization. Ensuring the ethical conduct of our research is of paramount importance, given the sensitive nature of the data and the potential implications for privacy. 
\begin{itemize}[leftmargin=*]
    \item \textbf {Data Collection and Usage:} Data was obtained from publicly available sources or with explicit permission, and sensitive information was excluded from our dataset. We respect any limitations set by data sources concerning data usage. 
    \item \textbf{Transparency and Explainability:} We recognize the importance of providing transparent and understandable summaries. Efforts were made to ensure that our summarization process can be explained to users, and address any algorithmic biases that may arise, striving for fairness.
    \item \textbf{Fairness and Bias Mitigation:} We are mindful of potential biases and employ measures to mitigate them, particularly those related to gender, race, and other sensitive attributes. Fairness in our summaries is a core consideration. 
    \item \textbf{Compliance with Regulations:} Our research adhered to all relevant data protection laws and regulations, such as GDPR, HIPAA, or regional data privacy laws, to ensure the ethical use of data.
    \item \textbf{Potential Risks and Mitigation Strategies:} We acknowledge the potential risks, such as unintended disclosure of sensitive information. To mitigate these risks, we implement strict controls and perform thorough risk assessments.
\end{itemize}
By taking these ethical considerations into account, we aimed to conduct our research on privacy policy text summarization with the utmost respect for privacy, transparency, and ethical integrity.

\section*{Acknowledgement}
Authors acknowledge the partial support of Infosys foundation through Center for AI (CAI)-IIIT Delhi.

\section*{Bibliographical References}\label{sec:reference}

\bibliographystyle{lrec-coling2024-natbib}
\bibliography{lrec-coling2024-example}

\end{document}